%% file: iclr2027_conference.tex
\documentclass{article} % For LaTeX2e
\usepackage{iclr2027_conference,times}

\input{math_commands.tex}

\usepackage[table]{xcolor}
\usepackage{hyperref}
\usepackage{cleveref}
\usepackage{tocloft}
\usepackage{url}
\usepackage{multirow}
\usepackage{tcolorbox}
\usepackage{mathtools}
\usepackage{booktabs}
\usepackage{tabularx}
\usepackage{array}
\usepackage{tikz}
\usepackage{caption}
\usepackage{graphicx}
\usepackage{subcaption}
\usepackage{listings}
\usepackage{xcolor}
\usepackage{arydshln}
\usepackage{algorithm}
\usepackage{algorithmicx}
\usepackage{wrapfig}
\usepackage{pifont}

\definecolor{mygreen}{rgb}{0.0, 0.5, 0.0}

\usepackage{algpseudocode}
\hypersetup{
    colorlinks=true,
    linkcolor=blue!50,
    urlcolor=blue!50,
    citecolor=blue!50
}

\lstdefinestyle{mypython}{
  language=Python,
  basicstyle=\ttfamily\scriptsize,
  backgroundcolor=\color{gray!5},
  keywordstyle=\color{blue},
  stringstyle=\color{green!60!black},
  commentstyle=\color{purple!80},
  numbers=left,
  numberstyle={\tiny\color{gray!90}},
  stepnumber=1,
  numbersep=8pt,
  frame=single,
  breaklines=true,
  tabsize=2,
  showstringspaces=false
}

\tcbset{
  promptstyle/.style={
    colback=brown!10,
    colframe=brown!50!black,
    fonttitle=\bfseries,
    coltitle=white,
    colbacktitle=brown!50!black,
    rounded corners,
    boxrule=0.75mm,
    coltext=black,
    width=\columnwidth,
    fontupper=\small
  }
}
\tcbset{
  casestyle/.style={
    colback=purple!10,
    colframe=purple!50!black,
    fonttitle=\bfseries,
    coltitle=white,
    colbacktitle=purple!50!black,
    rounded corners,
    boxrule=0.75mm,
    coltext=black,
    width=\columnwidth,
    fontupper=\small
  }
}

\title{Cliff: Learning Process Rewards from the First Mistake}

\author{
\hspace*{-0.1in}%
Peixuan Han$^{1,2,*}$
\hspace{-0.2em}
Runhui Wang$^{1}$
\hspace{-0.2em}
Ketan Ramaneti$^{1}$
\hspace{-0.2em}
Jie Hao$^{1}$
\hspace{-0.2em}
Gerald Friedland$^{1}$
\hspace{-0.2em}
Chris Kong$^{1,\dagger}$
\\
$^{1}$Amazon Web Services
\qquad
$^{2}$University of Illinois Urbana-Champaign
}

\NewDocumentCommand{\peixuan}
{ mO{} }{\textcolor{purple}{\textsuperscript{\textit{peixuan}}\textsf{\textbf{\small[#1]}}}}

\NewDocumentCommand{\chris}
{ mO{} }{\textcolor{orange}{\textsuperscript{\textit{chris}}\textsf{\textbf{\small[#1]}}}}

\iclrfinalcopy % Uncomment for camera-ready version, but NOT for submission.

\begin{document}

\maketitle
\begingroup
\renewcommand{\thefootnote}{\fnsymbol{footnote}}
\footnotetext[2]{Correspondence to: Chris Kong, \texttt{luyankon@amazon.com}.}
\footnotetext[1]{Work done during an internship at Amazon Web Services.}
\endgroup

\input{sections/abstract}
\input{sections/introduction}
\input{sections/related_work}
\input{sections/method}
\input{sections/judge_quality}
\input{sections/experiment}

\input{sections/analysis}
\input{sections/conclusion}

\bibliography{iclr2027_conference}
\bibliographystyle{iclr2027_conference}

\input{sections/appendix}

\end{document}

%% file: math_commands.tex
\usepackage{amsmath,amsfonts,bm}

\def\eqref#1{equation~\ref{#1}}
\def\1{\bm{1}}

\DeclareMathAlphabet{\mathsfit}{\encodingdefault}{\sfdefault}{m}{sl}
\SetMathAlphabet{\mathsfit}{bold}{\encodingdefault}{\sfdefault}{bx}{n}

%% file: sections/abstract.tex
\begin{abstract}

Reinforcement learning with verifiable rewards (RLVR) has emerged as a powerful paradigm for large language model (LLM) post-training, but its reliance on coarse outcome rewards leads to limited guidance on intermediate reasoning processes.
Existing approaches such as process reward modeling and on-policy distillation introduce additional constraints, such as reliance on a specialized reward model or assuming identical reasoning patterns between teacher and student. Nevertheless, we observe that once a reasoning process first goes wrong, evaluating the subsequent reasoning provides limited additional information, as it is already conditioned on an invalid prefix. Therefore, we propose \textbf{Cliff}, a reward shaping strategy that utilizes an off-the-shelf LLM as a teacher to identify the first mistake in each rollout. As a result, the rollout is naturally decomposed into two parts: a correct prefix and an incorrect suffix. Cliff then converts this signal into token-level advantages, assigning positive advantages for the correct prefix and negative feedback afterward.
Experiments across 12 different scenarios demonstrate that Cliff consistently improves reasoning performance, outperforming on-policy distillation by 15\% and standard GRPO by 7\%, even with teachers of modest capability. Furthermore, we analyse the role of ``ground truth'' in Cliff and investigate its training dynamics. These results establish Cliff as a simple, general and effective approach for improving RLVR with richer, fine-grained supervision.

\end{abstract}

%% file: sections/introduction.tex
\section{Introduction}

Reinforcement learning with verifiable rewards (RLVR) has become the standard paradigm for post-training large language models (LLMs) on reasoning tasks. By leveraging automatically verifiable outcomes, RLVR enables scalable, unbiased reinforcement learning without human annotations and has demonstrated remarkable progress in various domains. Despite its success, the outcome reward used in RLVR is inherently coarse-grained as it only evaluates the final result of a reasoning trajectory. As a consequence, it cannot distinguish between reasoning processes of vastly different quality. For example, a nearly complete solution containing only a minor error is penalized identically to a completely incorrect attempt. Such sparse, result-oriented supervision provides limited guidance on where the model succeeds or fails during reasoning, reducing learning efficiency and making it difficult to assign credit to individual reasoning steps.

Prior work has explored several strategies to incorporate fine-grained supervision into reinforcement learning, where two prominent directions are Process Reward Models (PRMs) and On-Policy Distillation (OPD). However, these approaches have notable limitations as well. PRMs require training an additional reward model, making it less generalizable and susceptible to reward hacking~\citep{zheng2024processbench,tiwari2026reward}. OPD only achieves optimal performance when the teacher and student share similar reasoning patterns and are in the same family~\citep{fu2026revisiting}. These constraints limit their applicability in general RLVR settings.

To overcome these limitations, we ask a more fundamental question: \textit{how fine-grained does a process signal really need to be?} Our key insight is that process supervision does not necessarily need to precisely evaluate every intermediate step, or every token. Once a reasoning process goes wrong, evaluating the subsequent reasoning may provide limited additional information, as it is already conditioned on an invalid prefix. This intuition is inspired by the notion of \textbf{Vacuous Implication} in formal logic: when the antecedent $A$ is false, the implication ``$A \rightarrow B$'' is true regardless of $B$. While this is not a formal proof for our method, it motivates a simple design principle: \textit{we only need to identify where the reasoning first goes wrong, rather than precisely evaluate everything that follows}. Based on this principle, \textbf{Cliff} divides each rollout exactly once at its first mistake, yielding only two segments—a correct prefix and an incorrect suffix—and uses this coarse-grained structure to provide process supervision without requiring token- or step-level rewards.

Specifically, we utilize a teacher model to judge student rollouts and locate the first mistake in the student's reasoning. For a rollout that fails to reach the final answer, tokens before the first mistake receive relatively higher advantage scores, while tokens after the boundary receive negative feedback just like GRPO. This decomposition into a correct prefix and an incorrect suffix offers several desirable properties. First, Cliff retains the simple formulation of RLVR while providing more informative process-level feedback: instead of assigning a single outcome-level reward to the entire rollout, Cliff distinguishes the valid reasoning from the point at which it first becomes invalid. Second, Cliff is both model-agnostic and task-agnostic. The supervision depends only on whether the reasoning remains correct up to a given point, rather than on the domain or the teacher's reasoning patterns. Thus, it can be applied across different models and tasks, provided that an evaluator is capable of identifying the first mistake. Finally, this design is naturally resistant to reward hacking. Since the reward is determined by the location of the first mistake, and postponing a mistake is always preferable to making the same mistake earlier, while a completely correct rollout remains optimal by definition.

Empirically, we first show LLM teachers can judge student answers and find out the first problematic reasoning accurately in \Cref{sec:judge_quality}, even when the model's own performance isn't perfect. We then show that Cliff outperforms on-policy distillation by 15\% and GRPO by 7\% across 12 different scenarios in \Cref{sec:exp}. We also demonstrate that incorporating ground truth is useful for weaker judge models, and analyse the training dynamics of Cliff. Overall, Cliff provides a simple yet general framework for transforming coarse outcome-based supervision into informative learning signals, paving the way toward more scalable and reliable reinforcement learning for LLMs.

%% file: sections/related_work.tex
\section{Related Work}

\textbf{Reinforcement Learning with Verifiable Rewards (RLVR).}
Due to its effectiveness and scalability, reinforcement learning (RL) has become a powerful paradigm for LLM post-training for reasoning tasks~\citep{qian2026toolrl,han2025tomap,yu2025sotopia}, and have been widely used by state-of-the-art LLMs~\citep{guo2025deepseek,jaech2024openai,team2025kimi}. Common algorithms for RLVR includes Proximal Policy Optimization (PPO)~\citep{schulman2017proximal}, Group Relative Policy Optimization (GRPO)~\citep{shao2024deepseekmath} and Dynamic Sampling Policy Optimization~\citep{yu2026dapo}. Many recent work focuses on understanding~\cite{zeng2025simplerl,mroueh2025reinforcement,wang2026beyond} and improving these methods, namely Dr.GRPO~\citep{liu2025understanding}, Clip-Cov~\citep{cui2025entropy} and Self-aligned Reward~\citep{han2026selfaligned}.

\textbf{LLM Distillation and Reward Modeling.} 
Distillation and reward modeling are both knowledge transfer approaches for LLMs. Knowledge distillation typically trains a student model to imitate a teacher's behavior~\citep{kim2016sequence,jiao2020tinybert,wang2020minilm}. On-policy distillation (OPD), as a more recent approach, trains the student on its own trajectories. OPD has shown strong performance~\citep{agarwal2024policy,lu2025onpolicydistillation,zhao2026selfdistilledreasoneronpolicyselfdistillation,yang2026oprd} and attracts increasing analysis on mechanisms~\citep{li2026rethinking,song2026survey,yang2026learning}, but its requirement for identical tokenizer and reasoning patterns limit its broader application~\citep{li2026rethinking,fu2026revisiting,boizard2024towards}. 

Reward models (RMs) are a key component of LLM post-training and alignment. They can be trained to generate scalar scores~\citep{ouyang2022training,rafailov2024direct,li2024llmr} as well as textual judgments~\citep{zhao2026genprm,chen2025rm}. Although commonly applied to unverifiable tasks, RMs have also proven effective for reasoning tasks by introducing process reward~\citep{setlur2025rewarding,lightman2024let,wang2024math,song2025prmbench}. However, some recent work suggests that process RMs require extensive data to ensure generalizability~\citep{zheng2024processbench,zeng2025versaprm}, and often suffer from reward hacking~\citep{cheng2026stop,tiwari2026reward} or degeneration to outcome rewards~\citep{zhang2025lessons}.

% \textbf{Reward Shaping in LLM Post-Training.} Reward shaping extends standard rewards to give denser learning signals.

% Distribution-level distillation typically requires the student and teacher to share the same tokenizer, and overcoming this barrier has become a key research topic~\citep{boizard2024towards,patino2025unlocking,niu2026breaking}.  Meanwhile, researchers are also investigating the mechanism behind OPD~\citep{li2026rethinking,song2026survey}, indicating that OPD works best when the teacher and student are similar in reasoning style.

%% file: sections/method.tex
\begin{figure}[t]
    \centering
    \includegraphics[width=0.95\textwidth]
    {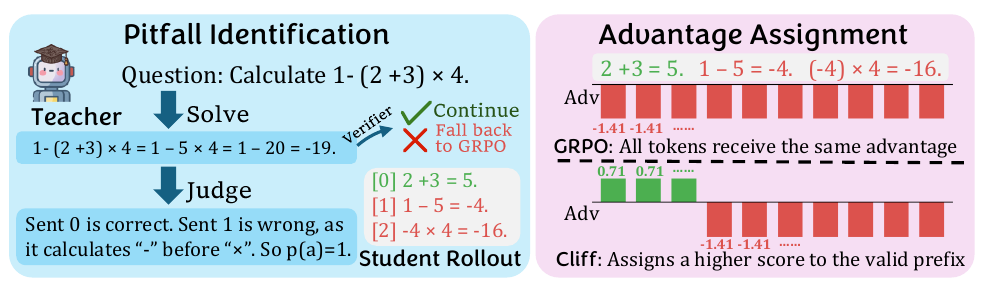}
    \vspace{-1em}
    \caption{Overview of \textbf{Cliff}.}
    \label{fig:main_fig}

\end{figure}

\section{Methodology}
\label{sec:method}

\subsection{GRPO}

Cliff is an extension based on Group Relative Policy Optimization (GRPO)~\citep{shao2024deepseekmath}, a widely adopted algorithm in reinforcement learning. In the GRPO formulation, we denote the current policy as $\pi_\theta$, the training set as $D$, and the reward function as $R$. For each step, we sample a query $q \sim D$, and several rollouts $a_{1:N}\sim \pi_{\theta_{\mathrm{old}}}(q)$. The objective function can be written as\footnote{In practice, we use DAPO's token-level loss aggregation~\citep{yu2026dapo}.}:

\begin{equation}
  \label{eq:grpo}
  \small
  \resizebox{\textwidth}{!}{$
  \mathcal{J}_{\text{GRPO}}(\pi_\theta) =
  \mathbb{E}
  \left[
  \frac{1}{\sum_{i=1}^{N}|a_i|}\sum_{i=1}^N\sum_{j=0}^{|a_i|-1}
  \left(
  \min\left\{
  r_{i,j} A_{i,j},
  \operatorname{clip}(r_{i,j}, 1-\epsilon, 1+\epsilon) A_{i,j}
  \right\}
  -\beta\, \mathbb{D}_{\text{KL}}\!\left[\pi_\theta \| \pi_{\text{ref}}\right]_{i,j}
  \right)
  \right].
  $}
\end{equation}   

\vspace{-0.5em}

In the formula, $r_{i,j} =
\frac{\pi_\theta(a_{i,j}\mid q,\, a_{i,<j})}
     {\pi_{\theta_{\mathrm{old}}}(a_{i,j}\mid q,\, a_{i,<j})}$ is the importance sampling ratio and $A_{i,j}$ is the advantage:

\vspace{-0.5em}

\begin{equation}
\small
\label{eq:advantage}
A_{i,j}=\frac{R(q,a_i)-\operatorname{mean}[R(q,a_1),...,R(q,a_N)]}{\operatorname{std}[R(q,a_1),...,R(q,a_N))]}.
\end{equation}

\vspace{-0.5em}
\subsection{Cliff}

To provide universal, fine-grained process supervision for RLVR, we propose \textbf{Cliff}, which identifies the first mistake in a student rollout and decomposes it into a correct prefix and an incorrect suffix. An overview of Cliff is illustrated in \Cref{fig:main_fig}.

\subsubsection{Pitfall Step Identification}

Cliff leverages a stronger teacher model to provide process-level feedback for each student rollout. The supervision is performed in two stages. In the first stage, the teacher independently generates a reference solution for each query. We then use the automatic verifier to evaluate the teacher's solution, and only keep groups where the teacher itself solves correctly. If the teacher's solution is incorrect, the corresponding group falls back to vanilla GRPO, as the teacher's guidance is no longer reliable. In the second stage, the teacher judges each student rollout based on the question and its own verified solution. The teacher first determines whether the student answer is correct and, if it is incorrect, identifies the first reasoning mistake. 

We refer to the first step at which the reasoning becomes incorrect as the \emph{Pitfall Step}, denoted by $p(a)$. The Pitfall Step serves as the boundary that separates each rollout into a correct prefix and an incorrect suffix. Thus, each incorrect rollout can be decomposed into a valid prefix up to $p(a)$ and a problematic suffix after $p(a)$.\footnote{For overlength rollouts, we directly set $\mathit{p}(a)=0$, treating the entire sequence as problematic. This is both for efficiency and to avoid length hacking (see \Cref{app:theory}).} We carefully curate the judging prompt to ensure the teacher focuses on genuine reasoning errors rather than harmless typos or different but valid solution strategies (prompts are shown in \Cref{app:prompt}).

\subsubsection{Cliff Advantage Assignment}

After obtaining the Pitfall Step $p(a)$, we convert the teacher's judgment into token-level advantages for GRPO. The key idea is to refine the outcome-level advantage of GRPO by distinguishing the valid prefix from the erroneous suffix within an incorrect rollout.

We first calculate the GRPO-style advantage. Without loss of generality, we assume that the reward is binary (i.e., $R(q,a) \in {0,1}$) \footnote{For domains with non-binary rewards (like coding as shown in later sections), we can either set a threshold or compare the average within a group again. Recent work~\citep{chen2025train,shu2026sailrlguidingmllmsthink,lee2026unified} shows that binary rewards could be more stable and less noisy in verifiable tasks, supporting this design.}, so the advantage values in GRPO within a group take only two possible values: $A_{cor}$ shared by all tokens in correct rollouts and $A_{inc}$ shared by all tokens in incorrect rollouts. Let the group size be $N$, and the group's reward statistics and advantages are therefore given by

\begin{equation}
\small
\label{eq:r}
\mu = \frac{\sum_{i \in 1:N}R(q,a_i)}{N}, \qquad
\sigma = \sqrt{\mu (1-\mu)}, \qquad
A_{\mathrm{cor}}
=
\frac{1-\mu}{\sigma},
\qquad
A_{\mathrm{inc}}
=
\frac{0-\mu}{\sigma}.
\end{equation}

We then utilize the \emph{Pitfall Step} to obtain the token-level advantages in Cliff. Intuitively, tokens before the \emph{Pitfall Step} $\mathit{p}(a)$ correspond to reasoning that remains valid, whereas tokens at and after the Pitfall Step are responsible for the erroneous reasoning and should receive negative feedback. Therefore, the token-level advantage in Cliff is defined as

We then use the Pitfall Step to refine the advantage within each rollout. For an incorrect rollout, tokens before the Pitfall Step $\mathit{p}(a)$ correspond to reasoning that remains valid, whereas tokens at and after the Pitfall Step belong to the reasoning that leads to the incorrect outcome. We therefore give a higher advantage for the valid prefix while assigning the negative outcome-level advantage to the erroneous suffix. The resulting token-level advantage in Cliff is defined as

\begin{equation}
\small
\label{eq:cliff_advantage}
A_{i,j} =
\begin{cases}
A_{\mathrm{cor}} - b,
& R(q,a_i)=1,\\

\lambda A_{\mathrm{cor}} - b,
& R(q,a_i)=0 \text{ and } j < \mathit{p}(a_i),\\

A_{\mathrm{inc}} - b,
& R(q,a_i)=0 \text{ and } j \ge \mathit{p}(a_i).
\end{cases}
\end{equation}

Here, $\lambda \geq 0$ is a hyperparameter that controls the strength of positive reinforcement assigned to the \emph{valid prefix} of an incorrect rollout. Empirically, we use $\lambda=0$ as it avoids length hacking (see \Cref{sec:lambda} and \Cref{app:theory}). The offset term $b$ is introduced to ensure that the token-level advantages have zero mean across the group. Specifically, the offset $b$ is equal to the average token-level advantage before this recentering operation:

\begin{equation}
\label{eq:offset}
\small
b=
\frac{
A_{\mathrm{cor}}
\sum_{i:\,R(q,a_i)=1}|a_i|
+
\sum_{i:\,R(q,a_i)=0}
\left[
\lambda A_{\mathrm{cor}}\mathit{p}(a_i)
+
A_{\mathrm{inc}}\bigl(|a_i|-\mathit{p}(a_i)\bigr)
\right]
}{
\sum_{i=1}^{N}|a_i|
}\footnotemark.
\end{equation}

\footnotetext{Token positions are indexed from $0$, so the valid prefix
$\{j < \mathit{p}(a_i)\}$ contains exactly $\mathit{p}(a_i)$ tokens.}

%% file: sections/judge_quality.tex
\section{Judge Quality of Cliff}
\label{sec:judge_quality}
This section conducts a preliminary experiment to evaluate whether the teacher model can reliably identify the ``Pitfall Step'' in student reasoning trajectories. We compare the teacher's judgments with human annotations from two perspectives: its ability to determine whether a student solution is correct and its ability to accurately localize the first reasoning mistake.

\subsection{Settings}
To facilitate a controlled evaluation, we construct a small, human-annotated dataset from DAPO-Math. We first sample a large number of student rollouts and then select 50 correct and 50 incorrect rollouts to form a balanced dataset. For each incorrect rollout, human experts are asked to identify the first reasoning step that leads to the incorrect conclusion (i.e., the Pitfall Step). We provide human annotators with the same instructions used by the LLM judge, ensuring that the two types of judgments are directly comparable.

\subsection{Judge Performance}

We evaluate the teacher models from three complementary perspectives. First, we measure each teacher's problem-solving accuracy by allowing it to solve the problems independently. Second, we evaluate its judging ability by asking it to determine whether each student solution is correct. Finally, for incorrect student solutions, we further compare the Pitfall Steps identified by the LLM judge and human experts. To quantify the alignment between human experts and the LLM judge, we define
\vspace{-0.5em}
\begin{equation}
\operatorname{p-dis}(a)=|p_{\text{human}}(a)-p_{\text{LLM}}(a)|,
\end{equation}
\vspace{-0.2em}

where $p_{\text{human}}(a)$ and $p_{\text{LLM}}(a)$ denote the sentence indices of the first incorrect reasoning step identified by the human annotator and the LLM, respectively. A smaller $\operatorname{p-dis}$ indicates closer agreement between the two, with a value of zero meaning that the LLM identifies exactly the same pitfall step as the human expert.

\begin{table}[t]
\caption{Teacher models' reference solution accuracy, judge consistency with the automatic verifier, and the similarity of Pitfall Step positions with human annotators.}
\vspace{-0.4em}
\label{table:judge_auality}
\small
\centering
\renewcommand{\arraystretch}{1.15}
\setlength{\tabcolsep}{4pt}

\resizebox{0.9\textwidth}{!}{%
\begin{tabular}{ll|c|ccc|cc}
\toprule
\textbf{Teacher Model} & \textbf{Ref. Solution}
& {Ref. Acc.} & {Judge Acc.} & {\# FP} & {\# FN}
& {Avg. $\operatorname{p-dis}$} & {$\operatorname{p-dis}\leq 1$} \\
\hline
Qwen3-32B
& \multirow{3}{*}{\shortstack{Provided by\\the Teacher}}
& 65 & 88 & 1 & 11 & 3.00 & 68 \\
Gemma3-27B
& & 67 & 86 & 0 & 14 & 4.52 & 39 \\
SOTA
& & 93 & 91 & 0 & 9 & 1.23 & 82 \\
\hline
Qwen3-32B
& \multirow{3}{*}{Ground Truth}
& \multirow{3}{*}{100} & 91 & 0 & 9 & 3.30 & 57 \\
Gemma3-27B
& & & 93 & 0 & 7 & 3.77 & 52 \\
SOTA
& & & 92 & 0 & 8 & 1.21 & 80 \\
\hline
Qwen3-32B
& \multirow{3}{*}{\shortstack{An Incorrect\\Solution}}
& \multirow{3}{*}{0} & 60 & 4 & 36 & 4.70 & 42 \\
Gemma3-27B
& & & 51 & 5 & 44 & 5.56 & 44 \\
SOTA
& & & 81 & 4 & 15 & 2.05 & 73 \\
\bottomrule
\end{tabular}
   
}
\end{table}

% From the first part of \Cref{table:judge_auality}, we can find that 

% 1.SOTA achieves the highest problem-solving accuracy and also produces the most reliable judgments, with its identified pitfall steps exhibiting strong agreement with human annotations. This suggests that highly capable models can not only solve mathematical problems effectively but also accurately localize the first reasoning error in incorrect solutions.

% 2. the ability to judge a solution is considerably more robust than the ability to generate one. Although Qwen3-32B and Gemma3-27B obtain noticeably lower solving accuracy than SOTA, they are still able to identify pitfall steps with high accuracy and maintain good agreement with human experts. This observation is encouraging for Cliff because it indicates that the teacher does not need to be a perfect problem solver. Instead, it only needs to reliably distinguish correct reasoning from incorrect reasoning and identify where the reasoning first deviates, a substantially easier task than solving the problem from scratch.

From the upper part of \Cref{table:judge_auality}, we can draw the following conclusions:

\textbullet \hspace{1pt} \textbf{Strong problem-solving ability leads to reliable process judgment.}
SOTA achieves the highest problem-solving accuracy and also produces the most reliable judgments, with its identified Pitfall Steps showing strong agreement with human annotations. This suggests that highly capable models can not only solve mathematical problems effectively, but also reliably localize the first reasoning error in incorrect student solutions.

\textbullet \hspace{1pt} \textbf{Judging is more robust than solving.} Although Qwen3-32B and Gemma3-27B obtain noticeably lower problem-solving accuracy than SOTA, they can still reliably identify Pitfall Steps and maintain good agreement with human experts. This observation is encouraging for Cliff, as it suggests identifying the mistake might be substantially easier than solving the problem from scratch.

\textbullet \hspace{1pt} \textbf{Most discrepancies come from false negatives.}
The judge rarely incorrectly rejects a correct student solution (i.e., false positives), while false negatives occur in approximately 10\% of cases. Importantly, some of these false negatives may not indicate genuinely unreliable judgment, as the automatic verifier may falsely accept solutions that arrive at the correct final answer through guessed answers or flawed reasoning.

\subsection{Impact of Reference Solution Quality}

This section analyzes how the quality of the reference solution affects the teacher's judgment. Specifically, we investigate two variants: in the first variant, the teacher is provided with a correct reference solution, while in the second variant, it is intentionally provided with an incorrect reference solution.

As shown in \Cref{table:judge_auality}, the quality of the reference solution has a clear impact on judgment performance. Providing a correct reference solution leads to higher judgment accuracy and more accurate localization of the Pitfall Step, whereas an incorrect reference solution degrades both metrics and results in a larger $\operatorname{p-dis}$.

This finding also motivates the reference-solution filtering procedure in Cliff. As described in \Cref{sec:method}, we use the automatic verifier to filter the teacher's reference solution and only proceed when the reference solution is verified as correct. Therefore, the ``Ground Truth'' setting provides a more faithful estimate of the judge's performance in actual Cliff training. Under this setting, the judgment accuracy exceeds $90\%$ and the average $\operatorname{p-dis}$ is only around 3 sentences, indicating that the teacher can reliably identify the first reasoning error when supplied with a verified reference solution. % This result further supports the reliability of Cliff's two-stage design, where reference-solution verification serves as a safeguard against propagating unreliable teacher judgments.

\subsection{Case Studies}

To better understand the behavior of Cliff, we present several representative examples in \Cref{app:cases} using Qwen3-32B as the teacher model. These examples provide qualitative evidence for the effectiveness of the proposed judging framework. We find that the teacher can analyze the student's reasoning step by step and accurately identify the earliest reasoning error in most cases, rather than simply checking whether the final answer is correct. In particular, the teacher is able to detect problematic reasoning even when a student arrives at the correct final answer, demonstrating that its judgment captures the validity of the reasoning process beyond the outcome alone. While the teacher's $p(a)$ does not always exactly match that of human annotators, which reflects the inherent ambiguity in localizing the first mistake, its decisions are generally reasonable. Overall, these examples suggest that Cliff can provide meaningful process-level supervision.

%% file: sections/experiment.tex
\section{Experiments}
\label{sec:exp}

\subsection{Settings}
\label{sec:exp_setting}

\textbf{Models.} We utilize two student models: Qwen3-4B-Base~\cite{yang2025qwen3} and Phi-4-mini-Instruct~\cite{abdin2024phi}. We choose three teacher models: a state-of-the-art LLM, Qwen3-32B, Gemma3-27B~\citep{gemmateam2025gemma3technicalreport}. We apply supervised finetuning (SFT) for Qwen3-4B-Base on OpenThoughts~\citep{guha2025openthoughts} before RL to provide instruction-following ability.

\textbf{Datasets.} We test Cliff on two domains: math reasoning and coding. For math reasoning, the models are trained on DAPO-math-17k-processed~\citep{yu2026dapo}; for coding, the models are trained on Deepcoder. Details of all training and evaluation datasets are listed in \Cref{app:datasets}. Notably, we use a binary reward for coding, only giving a score of 1 when the code passes all test cases.

\textbf{Baselines.} We compare Cliff against the following baselines:

\textbullet \hspace{1pt} GRPO: The prevalent RL algorithm, using an automatic verifier to provide sequence-level signals.

\textbullet \hspace{1pt} GRPO with Teacher: Use the teacher to judge whether the student rollout is correct, but the same advantage is applied to the entire rollout.

\textbullet \hspace{1pt} Distillation~\citep{kim2016sequence}: Sample responses from the teacher, and train student models in a supervised fine-tuning style.

\textbullet \hspace{1pt} On-policy Distillation (OPD): Use the teacher's distribution on student rollouts to provide token-level signals. Only applied for open-source teachers\footnote{We use the k1 estimator~\citep{lu2025onpolicydistillation} for same-family OPD and the method from \cite{niu2026breaking} for cross-tokenizer OPD.}.

\textbf{Training Details.} We show training details in \Cref{app:training}.

% \begin{tabular}{l|ccccc}
% \toprule
% Setting      & GSM8k & MATH-500 & DAPO-math & AIME & Avg. \\ \hline
% Qwen3-4B-SFT &       &          &           &      &      \\
% + GRPO       &       &          &           &      &      \\
%              &       &          &           &      &      \\
%              &       &          &           &      &     \\ \bottomrule
% \end{tabular}

% TODO: Consider Removing GSM8k as it's too easy and can't see the gap
\begin{table}[t]
\caption{\textbf{Cliff}'s performance on math and coding benchmarks.}
\vspace{-0.4em}
\label{table:result}
\small
\centering
\renewcommand{\arraystretch}{1.15}
\setlength{\tabcolsep}{4pt}

\resizebox{\textwidth}{!}{%
\begin{tabular}{l|c|ccccc|cccc}

\toprule

\multirow{2}{*}{Setting} & \multirow{2}{*}{Teacher}
& \multicolumn{5}{c|}{\textbf{Math Reasoning}}
& \multicolumn{4}{c}{\textbf{Algorithmic Coding}} \\

&
& GSM8k & MATH-500 & DAPO & AIME & Avg.
& CodeContests & LiveCode & DeepCoder & Avg. \\

\hline

Qwen3-4B
& \multirow{2}{*}{N/A}
& 84.53 & 71.00 & 29.40 & 17.69 & 50.66
& 14.25 & 20.29 & 10.60 & 15.05 \\

\ \ + GRPO
&
& 92.80 & 79.00 & 42.90 & 32.01 & 61.68
& 23.75 & 27.66 & 21.20 & 24.20 \\[0.1em]
\arrayrulecolor{gray}
\hdashline
\arrayrulecolor{black}

\ \ + Distill
& \multirow{3}{*}{SOTA}
& 90.978 & 74.40 & 38.40 & 30.54 & 58.58
& 21.25 & 25.37 & 15.20 & 20.61 \\

\ \ + GRPO
&
& 92.04 & 80.00 & 44.00 & 33.44 & 62.37
& 23.25 & 28.64 & 22.60 & 24.83 \\

\ \ + \textbf{Cliff}
&
& \textbf{93.17} & \textbf{83.20} & \textbf{49.30} & \textbf{36.98} & \textbf{65.66}
& 26.25 & 27.82 & 23.80 & \textbf{25.96} \\[0.1em]
\arrayrulecolor{gray}
\hdashline
\arrayrulecolor{black}

\ \ + Distill
& \multirow{4}{*}{Qwen}
& 92.722 & 75.4 & 38.2 & 27.44 & 58.44
& 22.00 & 26.35 & 20.60 & 22.98 \\

\ \ + OPD
&
& 91.21 & 77.20 & 37.90 & 26.37 & 58.17
& 19.25 & 25.70 & 16.20 & 20.38 \\

\ \ + GRPO
&
& 92.12 & 78.40 & 44.90 & 29.37 & 61.20
& 25.50 & 27.66 & 23.40 & 25.52 \\

\ \ + \textbf{Cliff}
&
& 92.12 & \textbf{80.80} & \textbf{48.90} & \textbf{36.65} & \textbf{64.62}
& 26.50 & 28.15 & 23.40 & \textbf{26.02} \\[0.1em]
\arrayrulecolor{gray}
\hdashline
\arrayrulecolor{black}

\ \ + Distill
& \multirow{4}{*}{Gemma}
& 92.722 & 72.00 & 33.10 & 25.83 & 55.91
& 25.00 & 27.82 & 20.40 & 24.41 \\

\ \ + OPD
&
& 91.13 & 74.00 & 33.20 & 22.62 & 55.24
& 18.50 & 26.64 & 18.80 & 21.31 \\

\ \ + GRPO
&
& 91.73 & 78.40 & 45.50 & 32.44 & 62.02
& 24.00 & 29.13 & 21.40 & 24.84 \\

\ \ + \textbf{Cliff}
&
& 92.72 & \textbf{80.60} & \textbf{47.50} & \textbf{33.97} & \textbf{63.70}
& 27.25 & 27.50 & 22.00 & \textbf{25.58} \\

\hline

Phi-4-Mini
& \multirow{2}{*}{N/A}
& 86.58 & 49.40 & 17.00 & 7.93 & 40.23
& 18.25 & 19.80 & 16.80 & 18.28 \\

\ \ + GRPO
&
& 88.93 & 66.40 & 29.00 & 14.79 & 49.78
& 20.00 & 23.57 & 21.60 & 21.72 \\[0.1em]
\arrayrulecolor{gray}
\hdashline
\arrayrulecolor{black}

\ \ + Distill
& \multirow{3}{*}{SOTA}
& 88.779 & 59.00 & 23.40 & 15.43 & 46.65
& 20.25 & 20.13 & 19.60 & 19.99 \\

\ \ + GRPO
&
& \textbf{89.39} & 66.60 & 28.00 & 15.43 & 49.86
& 24.50 & 25.53 & 24.80 & \textbf{24.94} \\

\ \ + \textbf{Cliff}
&
& 89.08 & \textbf{69.00} & \textbf{31.80} & \textbf{17.04} & \textbf{51.73}
& 23.75 & 26.84 & 24.20 & 24.93 \\[0.1em]
\arrayrulecolor{gray}
\hdashline
\arrayrulecolor{black}

\ \ + Distill
& \multirow{4}{*}{Qwen}
& 91.13 & 70.20 & 24.90 & 14.89 & 50.28
& 17.75 & 24.39 & 18.60 & 20.25 \\

\ \ + OPD
&
& 89.39 & 65.60 & 24.70 & 12.76 & 48.11
& 19.25 & 22.13 & 19.80 & 20.39 \\

\ \ + GRPO
&
& 87.14 & 68.40 & \textbf{29.30} & 16.29 & 50.18
& 23.25 & 24.55 & 20.20 & 22.67 \\

\ \ + \textbf{Cliff}
&
& \textbf{90.06} & \textbf{69.40} & 29.10 & \textbf{17.36} & \textbf{51.48}
& 24.25 & 24.88 & 21.60 & \textbf{23.58} \\[0.1em]
\arrayrulecolor{gray}
\hdashline
\arrayrulecolor{black}

\ \ + Distill
& \multirow{4}{*}{Gemma}
& 90.296 & 64.40 & 22.00 & 14.58 & 47.82
& 20.00 & 23.08 & 18.60 & 20.56 \\

\ \ + OPD
&
& 87.26 & 52.00 & 24.90 & 13.43 & 44.40
& 18.75 & 21.77 & 19.40 & 19.97 \\

\ \ + GRPO
&
& \textbf{90.42} & 65.60 & 28.00 & 15.51 & 49.88
& 23.50 & 21.73 & 20.80 & 22.01 \\

\ \ + \textbf{Cliff}
&
& 89.23 & \textbf{68.60} & \textbf{29.60} & \textbf{16.18} & \textbf{50.90}
& 24.25 & 23.90 & 21.60 & \textbf{23.25} \\

\bottomrule

\end{tabular}

}
\end{table}

\subsection{Main Results}
\label{sec:exp_setting}

The main experimental results are reported in \Cref{table:result}. We make the following observations:

\textbullet \hspace{1pt} \textbf{GRPO provides a stable baseline.}
GRPO achieves decent performance across all experimental settings, improving over the SFT-style distillation and OPD baselines in most cases.

\textbullet \hspace{1pt} \textbf{Cliff consistently outperforms all other methods.}
Cliff achieves the best performance under every evaluated setting. The improvements are observed across different student models and benchmark domains, demonstrating the effectiveness and generality of the proposed framework.

\textbullet \hspace{1pt} \textbf{Cliff is not overly dependent on a strong teacher.}
As expected, employing a stronger teacher generally leads to better student performance, with SOTA-based teachers consistently outperforming their open-source counterparts. However, the performance gap is relatively modest, suggesting that Cliff does not require a near-perfect teacher. Even moderately capable teacher models can identify reasoning pitfalls and provide useful training signals, making Cliff practical and scalable in settings where access to frontier models is limited.

\textbullet \hspace{1pt} \textbf{The gain comes from Cliff's credit assignment rather than merely introducing a teacher.}
We use ``GRPO with teacher'' as an ablation setting to isolate the benefit of using a teacher from that of Cliff's two-fold credit assignment. This setting yields only marginal improvements over the original GRPO objective, while Cliff produces substantially larger gains. This comparison highlights that the key contribution of Cliff is not merely using an LLM as a reward model, but explicitly identifying the Pitfall Step and assigning credit separately to the reasoning before and after the first mistake. Such fine-grained credit assignment provides substantially more informative supervision than treating the entire reasoning trajectory as a single unit.

%% file: sections/analysis.tex
\section{Analysis}

\subsection{Cliff without Ground Truth Filter}
In the main experiments, Cliff employs a filtering procedure in which the teacher's self-generated reference solution is verified by the automatic verifier. Cliff is applied only to questions for which the teacher produces a verified correct solution, and vanilla GRPO with feedback from the auto verifier is used otherwise. However, such ground-truth verification is not always available in real-world settings. A natural question is whether Cliff can improve reasoning even in scenarios without ground truth, so in this section, we conduct an experiment without the filtering step. In this scenario, all process-level feedback is derived solely from the teacher's own reference solution and its judgments.

\begin{wraptable}[19]{r}{0.65\textwidth}
\caption{Comparison of Cliff with and without ground truth.}
\vspace{-0.4em}
\label{table:gt}
\small
\centering
\renewcommand{\arraystretch}{1.15}
\setlength{\tabcolsep}{4pt}

\resizebox{0.65\textwidth}{!}{%
\begin{tabular}{l|cc|ccccc}
\toprule
Student
& Teacher
& Use GT
& GSM8k & MATH-500 & DAPO & AIME & Avg. \\
\hline
\multirow{7}{*}{Qwen3-4B} 
& GRPO & Yes
& 92.80 & 79.00 & 42.90 & 32.01 & 61.68 \\
& \multirow{2}{*}{SOTA} & Yes
& 93.17 & 83.20 & 49.30 & 36.98 & 65.66 \\
& & No
& 93.25 & 81.20 & 48.50 & 36.65 & 64.90 \\
& \multirow{2}{*}{Qwen} & Yes
& 92.12 & 80.80 & 48.90 & 36.65 & 64.62 \\
& & No
& 90.11 & 78.80 & 47.70 & 34.84 & 62.69 \\
& \multirow{2}{*}{Gemma} & Yes
& 92.72 & 80.60 & 47.50 & 33.97 & 63.70 \\
& & No
& 89.94 & 78.60 & 46.60 & 34.27 & 62.35 \\
\hline
\multirow{7}{*}{Phi-4-Mini}
& GRPO & Yes
& 88.93 & 66.40 & 29.00 & 14.79 & 49.78 \\
& \multirow{2}{*}{SOTA} & Yes
& 89.08 & 69.00 & 31.80 & 17.04 & 51.73 \\
& & No
& 89.91 & 71.20 & 27.90 & 16.82 & 51.46 \\
& \multirow{2}{*}{Qwen} & Yes
& 90.06 & 69.40 & 29.10 & 17.36 & 51.48 \\
& & No
& 89.39 & 67.20 & 27.40 & 14.88 & 49.72 \\
& \multirow{2}{*}{Gemma} & Yes
& 89.23 & 68.60 & 29.60 & 16.18 & 50.90 \\
& & No
& 87.80 & 64.10 & 26.90 & 14.41 & 48.30 \\
\bottomrule
\end{tabular}

}
\end{wraptable}

From \Cref{table:gt}, we can find that the frontier-model teacher achieves almost the same performance in settings with and without the ground truth verifier, indicating that its self-generated reference solutions are already sufficiently accurate and reliable. In contrast, weaker teacher models such as Qwen3-32B and Gemma3-27B experience a performance drop of around 2\% when the ground-truth filter is removed. For these teacher models, removing the ground-truth filter means some inaccurate reference solutions will be used to judge student rollouts, disturbing the student's training signals. This suggests that \textbf{the primary function of the ground-truth filter is to compensate for the reference solution quality of less capable teachers}, while a strong teacher can provide reliable guidance on its own in most cases.

Notably, even open-source teacher models without ground-truth filtering still often outperform vanilla GRPO. This is consistent with the finding in \Cref{sec:judge_quality} that judging student rollouts is easier than solving problems from scratch, and further demonstrates that \textbf{Cliff can provide useful process-level supervision even when reliable ground-truth solutions are unavailable.}

 \begin{wraptable}[11]{r}{0.7\textwidth}
\caption{Cliff with different $\lambda$ values.}
\vspace{-0.4em}
\label{table:lambda}
\small
\centering
\renewcommand{\arraystretch}{1.1}
\setlength{\tabcolsep}{4pt}

\resizebox{0.7\textwidth}{!}{%
\begin{tabular}{l|cccccc}
\toprule
Setting & GSM8K & MATH-500 & DAPO-math & AIME & Avg. Acc. & Avg. Len. \\ \hline
GRPO          & 92.80  & 79.00 & 42.90  & 32.01 & 61.68 & 1279 \\
$\lambda=0.0$ & \textbf{93.17} & \textbf{83.20} & \textbf{49.30} & \textbf{36.98} & 65.66 & 1506 \\
$\lambda=0.5$ & 92.873 & 82.8   & 47.2   & 35.798 & 64.67 & 1481 \\
$\lambda=1.0$ & 93.252 & 81.2   & 47.8   & 33.655 & 63.98 & 1959 \\
\bottomrule
\end{tabular}
}
\end{wraptable}

\subsection{The Choice of $\lambda$}
\label{sec:lambda}
In Cliff, the hyperparameter $\lambda$ controls the amount of positive advantage assigned to the correct prefix of an incorrect reasoning trajectory, which is specified in \Cref{eq:cliff_advantage}. To investigate the effect of $\lambda$, we conduct Cliff training with three different values.

From the results in \Cref{table:lambda}, we can observe that $\lambda=0$ provides the best performance with reasonable generation length, and $\lambda=0.5$ achieves comparable performance. Increasing $\lambda$ to 1.0, however, leads to substantially longer responses and a noticeable degradation in reasoning performance. We attribute this behavior to the potential incentive for length hacking introduced by a large $\lambda$. When the valid prefix of an incorrect rollout receives a positive advantage, a larger $\lambda$ provides a stronger incentive to preserve and extend the already-correct portion of the reasoning before reaching more difficult or error-prone steps. In the extreme case of $\lambda=1.0$, the model may benefit from producing longer correct prefixes even when the additional reasoning is uninformative, thereby increasing response length without improving the final outcome. This lengthening effect can also make the model less effective at progressing toward the actual solution. 

In conclusion, \textbf{$\lambda=0$ leads to the most preferable results and is therefore used in our main experiments.} Specifically, $\lambda=0$ means the correct prefix of an incorrect rollout receives an advantage of $A=-b$, without additional positive reinforcement. The theoretical analysis in \Cref{app:theory} provides a formal characterization of this potential length-hacking behavior with similar conclusions, showing that $\lambda=0$ is free from length hacking, while a sufficiently large $\lambda$ value may promote longer but meaningless rollouts.

% Throughout the main experiments, we set $\lambda=0$, meaning that the correct prefix receives neither additional encouragement nor penalty. This design focuses the optimization on discouraging reasoning after the pitfall step while avoiding unnecessary optimization of already-correct prefixes.

\subsection{Training Dynamics}
\label{sec:curves}

\begin{figure}[t]
    \centering
    \includegraphics[width=0.32\textwidth]{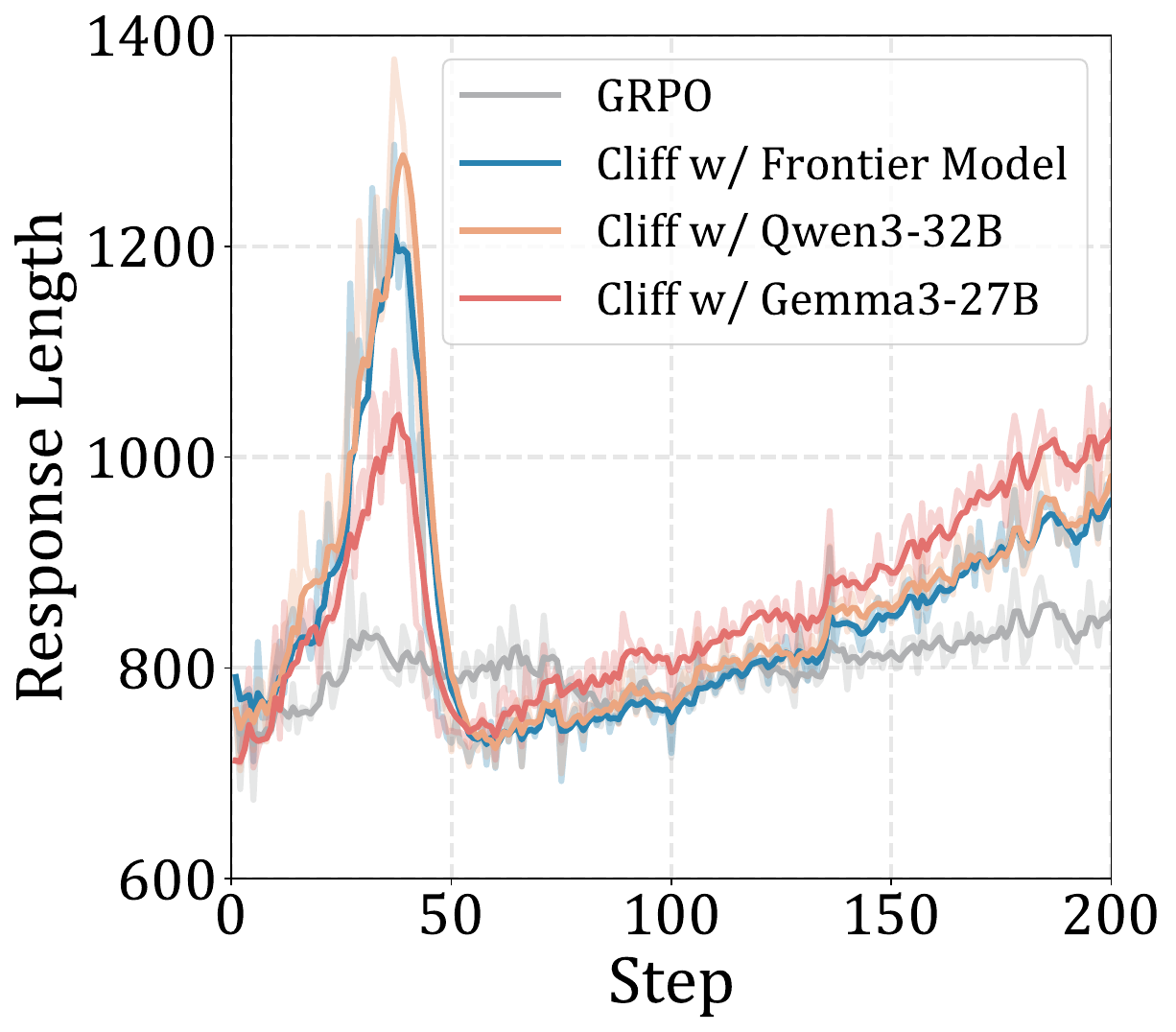}
    \includegraphics[width=0.32\textwidth]{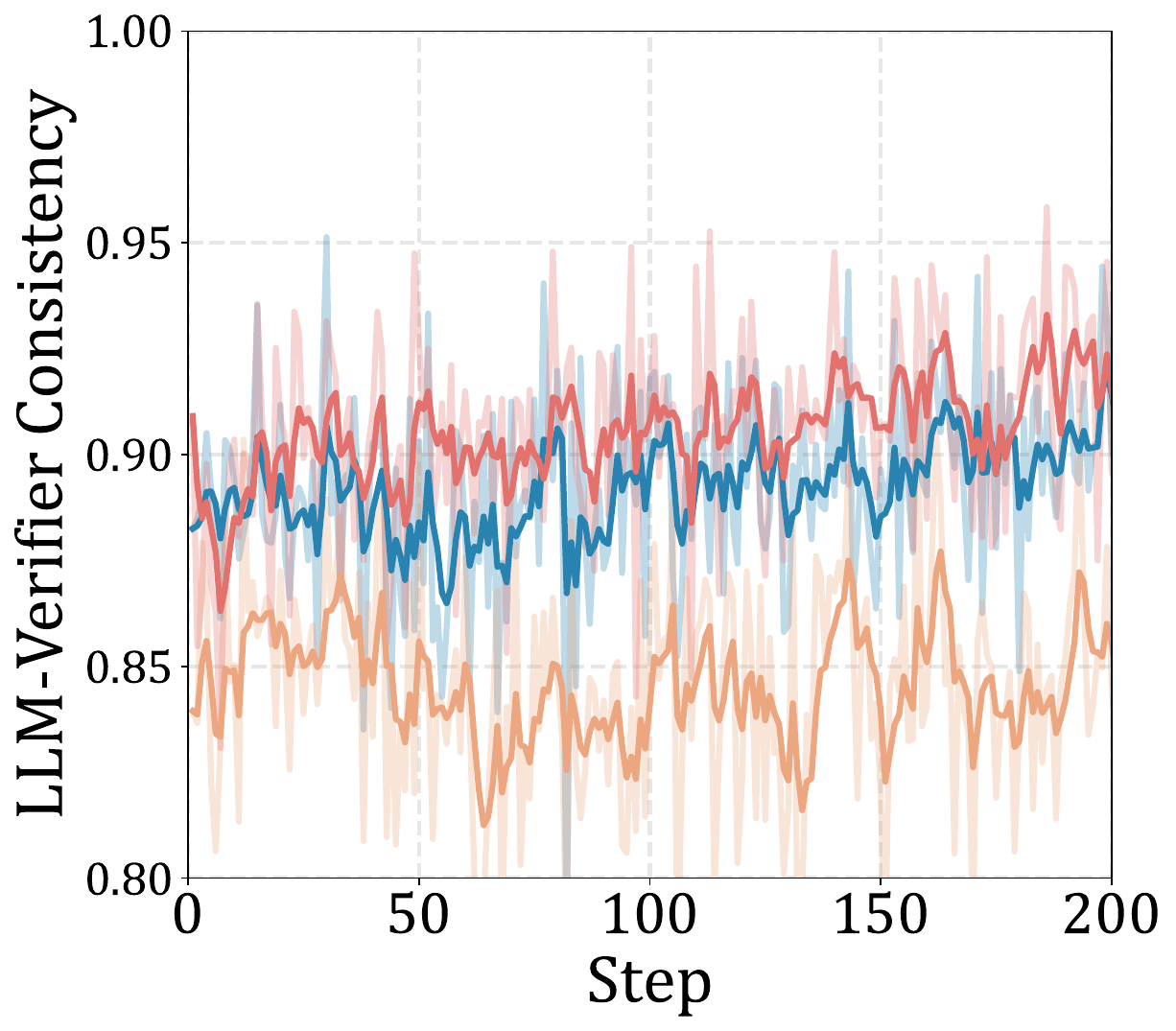}
    \includegraphics[width=0.32\textwidth]{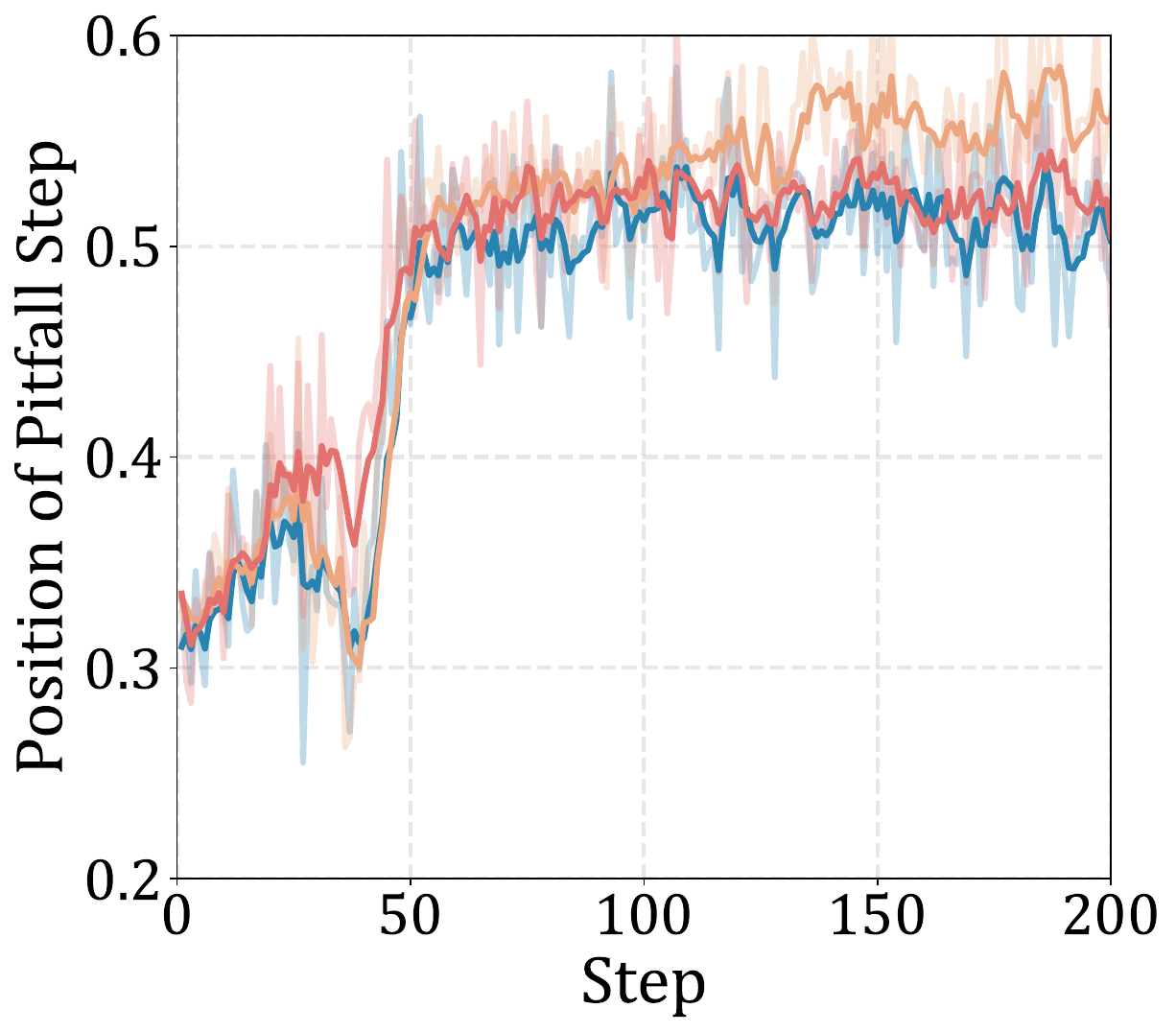}
    % \vspace{-2em}
    \caption{Training plots of GRPO and Cliff with different teachers\protect\footnotemark. The student model is Phi-4-Mini-Instruct trained on the math domain.}
    \label{fig:curves}

\end{figure}

\footnotetext{Consistency refers to the fraction of rollouts' outcome rewards judged the same by the teacher model and automatic verifier. The position of Pitfall Step means the average fraction of tokens in the valid prefix (before the first mistake) of an incorrect rollout.}

In this section, we analyze the evolution of different metrics throughout the training process to better understand the training dynamics of Cliff. The metrics are illustrated in \Cref{fig:curves}.

First, we examine the trend of response length during training. All settings gradually develop longer reasoning trajectories, indicating that the models learn to conduct deeper reasoning to math problems. Cliff generally produces longer responses than vanilla GRPO, with a noticeable spike in response length during the early stage of training. This suggests that Cliff initially encourages the model to explore longer reasoning trajectories before the generation length gradually stabilizes.

Second, we examine the consistency between teacher judgments and the automatic verifier. While the consistency is pretty high (85\% to 90\%), there are a non-negligible fraction of responses judged negatively by the teacher while being accepted by the verifier. This discrepancy is expected because the verifier primarily evaluates the final answer, whereas the teacher also considers the validity of the reasoning process. We also observe lower consistency for the Gemma teacher, indicating its relatively limited capability in reliably judging student reasoning.

Finally, we track the relative position of the Pitfall Step in the whole answer for incorrect rollouts. We observe that this ratio drastically increases in around 50 steps to 0.5, and then gradually stabilizes. This suggests that, early in training, the model learns to complete early steps of reasoning even for difficult problems, leading to relatively longer valid prefixes. As training progresses, the metric reaches a stable level, indicating that the model has established a more consistent reasoning process instead of hacking the teacher model. These dynamics suggest that Cliff encourages improvements in intermediate reasoning rather than merely optimizing the final answer.

%% file: sections/conclusion.tex
\section{Conclusion}
In this work, we introduce \textbf{Cliff}, a simple yet effective reward shaping algorithm that identifies the first mistake (Pitfall Step) in a rollout and assigns different advantages to tokens before and after it. Unlike approaches that rely on additional reward models or restrictive assumptions about reasoning trajectories, Cliff provides a task-agnostic and model-agnostic learning signal that works with teachers of different capabilities, bringing 7\% performance gain over vanilla GRPO on average. We also analyse the impact of ground truth and the training dynamics of Cliff.
Overall, this work proposes a direction to improve RLVR of LLMs with better ways of interpreting and utilizing the simple signals, instead of incorporating complex supervision mechanisms. In the future, we plan to extend the method to agentic settings and explore rule-based Pitfall Step detectors.

% \section*{AI use statement}

% In this work, we used generative AI tools for implement methods, assist with translation.
% We have not used generative AI tools for help develop theoretical models or conceptual frameworks, formulate mathematical claims, provide critical ingredients for proving mathematical claims, assist in the writing of proofs, propose or refine hypotheses, design or provide feedback on research methodology or experiments, clean and reformat dataset, support qualitative and thematic data analysis, interpret results. Additionally, we used generative AI tools for edit a research paper to improve readability, identify relevant literature. We have reviewed and humanly curated all AI-assisted work, including code, paper content and relevant work. We take responsibility for the final content of this work, including text, claims or artifacts produced with the aid of generative AI.

% \section*{Reproducibility Statement}

%% file: sections/appendix.tex
\newpage
\appendix
\renewcommand\thesection{\Alph{section}}
\renewcommand\thesubsection{\thesection.\arabic{subsection}}
\renewcommand\thesubsubsection{\thesubsection.\arabic{subsubsection}}

\section{Datasets}
\label{app:datasets}

For math datasets, we use no system prompt and only the raw question as the user prompt. During evaluation, we parse the last integer, fraction number, or content wrapped in \textit{\\boxed\{...\}} in the model's output and compare it with the ground truth.

For coding datasets, we also use no system prompt. We combine the raw question, an input-output example, and coding instructions in the user prompt. Due to resource limitations, we consider the single-turn coding task, where the model can't run its code in advance. During evaluation, we extract the code block in the model's answer and use the sandbox from VolcEngine\footnote{\url{https://hub.docker.com/r/volcengine/sandbox-fusion}} to test the code. We use 10 cases per question, set a time limit of 5 seconds, and only give a score of 1 when the model passes all test cases.

\Cref{table:datasets} shows the statistics of all datasets used. We show an example of each domain below.

\begin{table}[h]
\caption{Statistics of datasets.}
\label{table:datasets}
\small
\centering
\renewcommand{\arraystretch}{1.1}
\setlength{\tabcolsep}{4pt} % Reduces space to 4pt
\resizebox{\textwidth}{!}{%
\begin{tabular}{l|cccccccc}
\toprule
Dataset     & GSM8k & MATH-500 & DAPO & AIME & CodeContests & LiveCodeBench & DeepCoder & OpenThoughts \\\hline
License & MIT & MIT & Apache-2.0 & Apache-2.0 & CC BY 4.0 & CC & MIT & Apache-2.0 \\     
\# Train    & 7473  & -     & 13,116 & -    & -            & -        & 13.485  &  113,957      \\
\# Eval     & 1319  & 500     & 1000 & 933  & 400          & 611      & 500  & -     \\
\bottomrule
\end{tabular} }
\end{table}

\begin{center}
\begin{tcolorbox}[casestyle, title={An Example from DAPO-Math}]
The vertices of a regular nonagon ($9$-sided polygon) are to be labeled with the digits $1$ through $9$ in such a way that the sum of the numbers on every three consecutive vertices is a multiple of $3$. Two acceptable arrangements are considered to be indistinguishable if one can be obtained from the other by rotating the nonagon in the plane. Find the number of distinguishable acceptable arrangements.
\end{tcolorbox}
\end{center}

\begin{center}
\begin{tcolorbox}[casestyle, title={An Example from DeepCoder}]
Given  a permutation $P_1, \ldots, P_N$ of $1, \ldots, N$, find the number of integers $i$ ($1 \le i \le N$) that satisfy the following condition: for every integer $j$ ($1 \le j \le i$), $P_i \le P_j$.
\\

Constraints

- $1 \le N \le 2 \times 10^5$

- $P_1, \ldots, P_N$ is a permutation of $1, \ldots, N$.

- All values in the input are integers.
\\

Input

The input is given from standard input in the following format:

N

P1 P2 ... PN
\\

Output

Print the number of integers $i$ that satisfy the condition.
\\

Sample Input

5

4 2 5 1 3
\\

Sample Output

3
\\

% Explanation

% The indices $i=1$, $2$, and $4$ satisfy the condition.

% For $i=3$, the condition fails because $P_3 > P_1$.

% Similarly, $i=5$ does not satisfy the condition. Therefore, there are three valid indices.
% \\

Read the input from standard input (stdin) and write the answer to standard output (stdout). First reason step by step about your approach, then put your final solution in a single python code block. Your code should finish in 5s under the data size specified in the question.
\end{tcolorbox}
\end{center}

\section{Training Details}
\label{app:training}

\begin{table*}[t]
\caption{Training configurations.}
\label{table:hparams}
\small
\centering
\renewcommand{\arraystretch}{1.1}

\begin{minipage}[t]{0.31\textwidth}
\centering
\subcaption{Distill}
\begin{tabular}{@{}lc@{}}
\toprule
Hyperparameter & Value \\\hline
Actor learning rate & $1e-5$ \\
Batch Size & 384 \\
Training step & 400 \\
Max response length & 4096 \\
\bottomrule
\end{tabular}
\end{minipage}
\hfill
\begin{minipage}[t]{0.31\textwidth}
\centering
\subcaption{OPD}
\begin{tabular}{@{}lc@{}}
\toprule
Hyperparameter & Value \\\hline
Learning rate & $1e-6$ \\
Batch Size  & 160 \\
Training step & 1000 \\
Max response length & 4096 \\
Rollout temp ($\tau$) & 1.0 \\
Advantage clip ($\epsilon$) & 0.2 \\
Log Prob Clamp & -10 \\
KL Estimator & k1 \\
\bottomrule
\end{tabular}
\end{minipage}
\hfill
\begin{minipage}[t]{0.31\textwidth}
\centering
\subcaption{GRPO \& Cliff}
\begin{tabular}{@{}lc@{}}
\toprule
Hyperparameter & Value \\\hline
Actor learning rate & $1e-6$ \\
Batch Size & 64 \\
Num of rollouts & 12 \\
Training step & 200 \\
Max response length & 4096 \\
Rollout temp ($\tau$) & 1.0 \\
KL penalty ($\beta$) & 0 \\
Advantage clip ($\epsilon$) & 0.2 \\
\bottomrule
\end{tabular}
\end{minipage}

\end{table*}

We develop our code based on veRL~\citep{sheng2025hybridflow}. We provide key hyperparameters of all training algorithms including distill with teacher rollouts, on-policy distillation, and GRPO in \Cref{table:hparams}. For the warm-up SFT for Qwen3, we use the same setting as Distill, but train 2 epoch on the Openthought dataset. We follow standard practices in veRL and ensure different settings have comparable data budgets. Models are trained on 4 NVIDIA H100 80GB GPUs or B200 192GB GPUs\footnote{We find the vLLM inference engine may behave differently for Phi-4-mini on different GPUs, so Phi-4-mini experiments are conducted on H100s. Qwen-4-3B experiments are conducted on B200s.}.

For Cliff-specific parameters, we use $\lambda=0$ and a teacher rollout temperature of $0.6$. To save compute, we skip teacher judging for prompt groups that the automatic verifier scores unanimously — all rollouts correct or all incorrect. Such groups carry no reward variance and thus receive zero advantage under the GRPO-style loss. Skipping them only causes signal loss when the judge disagrees with the verifier, which is very rare on these groups where the question is too simple or too hard. During inference, we use greedy decoding and a maximum response length of 8192.

\section{Length Dynamics of Cliff}
\label{app:theory}

This section provides a theoretical analysis on length dynamics of Cliff. To summarize:

\textbullet \hspace{1pt} GRPO encourages lengthening when and only when correct rollouts are longer than incorrect ones.

\textbullet \hspace{1pt} Cliff is more ``lenient'' on lengthening than GRPO, and a larger $\lambda$ makes lengthening easier.

\textbullet \hspace{1pt} $\lambda=0$ eliminates the possibility to hack the reward by lengthening alone. $\lambda<\frac{L_{cor}}{L_{max}}$ is a sufficient in most conditions as it ensures the advantage of an incorrect rollout never exceed a correct one.

\textbullet \hspace{1pt} Hard capping $p(a)=0$ for overlength rollouts (considering the whole rollout as wrong) is necessary to stable training.

\subsection{Notations}

We follow the notation in \Cref{sec:method} and add the quantities of
Table~\ref{tab:notation}.

\begin{table}[htbp]
\centering
\small
\caption{Additional notations for the math analysis. % $\Phi$, $M$, $\Delta$ and $b$ arebper-group quantities. The length mass is written $\Phi$ because $D$ already denotes the training set.
}
\label{tab:notation}
\renewcommand{\arraystretch}{1.15}
\begin{tabular}{ll}
\hline
Symbol & Meaning \\
\hline
$L_{\max}$ & The maximum response length \\
$T = \sum_{i=1}^{N}|a_i|$ & The total number of response tokens in a group \\
$s_i = \mathit{p}(a_i)/|a_i|$ & The fraction of the correct prefix in an incorrect rollout \\
$N_{\mathrm{cor}}$, $N_{\mathrm{inc}}$ & The number of correct / incorrect rollouts in a group \\
$L_{\mathrm{cor}}$, $L_{\mathrm{inc}}$ & The average of $|a_i|$ over the correct / incorrect rollouts \\
$\rho = L_{\mathrm{cor}}/L_{\mathrm{inc}}$ & The ratio of correct and incorrect rollout lengths in a group \\
% $\Phi$ & \emph{length mass}: net lengthening pressure of the group \\
% $M$ & total advantage mass of the group ($M_0$ before the offset $b$ is applied) \\
% $\Delta$ & advantage mass injected by the prefix credit of \eqref{eq:cliff_advantage} \\
\hline
\end{tabular}
\end{table}

% Three conventions make the analysis exact, and they are the ones our implementation uses.
% (i) \emph{Token-level aggregation}: the outer normaliser of \eqref{eq:grpo} is
% $1/\sum_{i}|a_i|$, so every token of the batch carries the same weight.\footnote{Written with
% the per-sequence normaliser $\frac{1}{N}\sum_i\frac{1}{|a_i|}$ instead, the objective acquires a
% second, unconditional lengthening term driven by the harmonic rather than the arithmetic means of
% the two length classes, because a long incorrect rollout then absorbs a $1/|a_i|$ share of the
% penalty per token.} (ii) The KL coefficient is $\beta = 0$ and there is no entropy bonus, so the
% advantage is the only channel through which length can be affected. (iii) We work at first order
% in the initial inner epoch, where $r_{i,j} = 1$ and the clipping in \eqref{eq:grpo} is inactive,
% so the update direction is
% $\Delta\theta \propto \sum_{i,j} A_{i,j}\,\nabla_\theta \log \pi_\theta(a_{i,j}\mid q, a_{i,<j})$.
% (The sample-standard-deviation correction used in the implementation rescales the global step
% size only and changes no sign and no threshold below, so we write $\sigma$ as in \eqref{eq:r}.)

\subsection{Length Mass ($\Phi$) and Length Bias of GRPO}

% GRPO assigns $A_{i,j} = A_i$ to every token $j$ of rollout $i$, with
% $A_i = A_{\mathrm{cor}}$ if $R(q,a_i) = 1$ and $A_i = A_{\mathrm{inc}}$ otherwise.

The length of an autoregressive LLM policy's samples is controlled by the EOS probability at each
position. Raising a non-EOS token at a position where EOS was a live alternative lowers the
relative mass of EOS; raising
the EOS token itself shortens. \textbf{In other words, each untruncated rollout $a_i$ contributes $|a_i| - 1$ non-EOS tokens and exactly
one EOS token}. Given that GRPO assigns the same advantage to every token in a rollout, a group's net lengthening pressure (\emph{length mass}) can be defined as
\begin{equation}
\small
\Phi \;=\; \sum_{i=1}^{N}\Bigl[\sum_{j=0}^{|a_i|-2}A_{ij}-A_{i,|a_i|-1}\Bigr] =
  \;=\; \sum_{i=1}^{N} A_i |a_i| \;-\; 2\sum_{i=1}^{N} A_i
  \;=\; \sum_{i=1}^{N} A_i |a_i|.
\label{eq:phidef}
\end{equation}
The final equality holds because the mean-centering feature in GRPO (\label{eq:r}) gives
$\sum_{i=1}^{N} A_i = 0$.

By decomposing $A_{ij}$ to the correct tokens and incorrect tokens, we have:

\begin{equation}
\small
\begin{aligned}
\Phi &\;=\; \sum_{i=1}^{N} A_i |a_i|
\;=\; A_{\mathrm{cor}}\!\!\sum_{i|R(q,a_i)=1}\!\!|a_i|
 \;+\; A_{\mathrm{inc}}\!\!\sum_{i|R(q,a_i)=0}\!\!|a_i|
 \;=\; N\Bigl[\mu A_{\mathrm{cor}} L_{\mathrm{cor}} + (1-\mu) A_{\mathrm{inc}} L_{\mathrm{inc}}\Bigr]\\[2pt]
&=N\Bigl[\mu \frac{1-\mu}{\sigma} L_{\mathrm{cor}} - (1-\mu) \frac{\mu}{\sigma} L_{\mathrm{inc}}\Bigr]
\;=\; \frac{N\mu(1-\mu)}{\sigma}\,\bigl(L_{\mathrm{cor}} - L_{\mathrm{inc}}\bigr)
 \;=\; N\,\sigma\,\bigl(L_{\mathrm{cor}} - L_{\mathrm{inc}}\bigr).
\end{aligned}
\label{eq:phi}
\end{equation}

As a conclusion, GRPO's length mass $\Phi$ has the same sign of  $(L_{\mathrm{cor}} - L_{\mathrm{inc}})$, or $\Phi>0 \Longleftrightarrow \rho>1$. Therefore, \textbf{GRPO is a faithful amplifier of whatever correctness-length correlation the policy's own samples
happen to carry.} There is therefore no length channel to exploit -- lengthening cannot increase the objective unless longer responses are genuinely more often correct -- so reward hacking on
length does not occur.

\subsection{Length Mass of Cliff}

Since Cliff recenters token-level advantage to zero, we have $\sum_i\sum_jA_{ij}=0$. Therefore, the length mass of Cliff can be written as:

\begin{equation}
\small
\Phi \;=\; \sum_{i=1}^{N}\Bigl[\sum_{j=0}^{|a_i|-2}A_{ij}-A_{i,|a_i|-1}\Bigr] =
\sum_{i=1}^{N}\Bigl[\sum_{j=0}^{|a_i|-1}A_{ij}-2A_{i,|a_i|-1}\Bigr] = -2\sum_{i=1}^{N}A_{i,|a_i|-1} = 2Nb.
\label{eq:phicliff}
\end{equation}

The final equality holds because the advantage of the last token in a rollout is always $A_{\mathrm{cor}}-b$ for correct rollouts and $A_{\mathrm{inc}}-b$ for incorrect ones, since $p(a)\leq |a_i|-1$. Given that the outcome-level GRPO advantages are normalized within each group, $N_{\mathrm{cor}}A_{\mathrm{cor}}+N_{\mathrm{inc}}A_{\mathrm{inc}}=0$.

Using \Cref{eq:offset}, we can further derive:

\begin{equation}
\small
\begin{aligned}
\Phi &= 2Nb = \frac{2N}{T}\Bigl[A_{cor}(N_{cor}L_{cor}+N_{inc}L_{inc}\lambda \bar{s})+A_{inc}N_{inc}L_{inc}(1-\bar{s})\Bigr]\\
&=\frac{2N^2(1-\mu)L_{inc}}{\sigma T}\Bigl[\mu \rho + (1-\mu)\lambda \bar{s} - \mu (1-\bar{s})\Bigr]
\end{aligned}
\label{eq:phicliff_cont}
\end{equation}

Therefore:
\begin{equation}
\small
\Phi>0 \Longleftrightarrow \mu \rho + (1-\mu)\lambda \bar{s} - \mu (1-\bar{s})>0 \Longleftrightarrow \rho>1-\bar{s}(1+\lambda\frac{1-\mu}{\mu}).
\label{eq:phicliff_euivalent}
\end{equation}

From this formulation, we can find that:

\textbullet \hspace{1pt} If $\bar{s}=0$, then Cliff degenerates to GRPO, where the lengthening condition is $\rho>1$.

\textbullet \hspace{1pt} When $\bar{s}>0$, we denote $f(\lambda)=1+\lambda\frac{1-\mu}{\mu}$, then we have $f(\lambda)>0$ and $df(\lambda)/d\lambda>0$. This means \textbf{Cliff encourages lengthening more than GRPO does (the lengthening condition is easier to achieve), and a larger $\lambda$ makes the lengthening trend stronger.}

\paragraph{The importance of truncation.} Previous analysis is built upon an assumption that the EOS token appears in each rollout. This assumption might not hold when some rollouts reach $L_{max}$ and are truncated. In GRPO, overlength rollouts almost always receive a zero reward and negative advantage, suppressing the lengthing trend. In Cliff, however, the teacher may give a long valid prefix for reasoning that is not wrong but just did not finish. That's why we set a hard cap $\mathit{p}(a) = 0$ for overlength rollouts in Cliff to ensure overlength rollouts are penalized. This setting is thus not merely a saving of judge calls but a necessary guard.

\subsection{Cliff's Robustness to Reward Hacking}

We consider the possibility of length reward hacking when $b>0$, or the overall trend of a group is lengthening. Length hacking is excluded if two conditions hold: \textbf{(C1)} a correct rollout always outranks
an incorrect one in total advantage, and \textbf{(C2)} an incorrect rollout cannot raise its total
advantage by lengthening its valid prefix.

For \textbf{(C1)}:

\begin{equation}
\small
\begin{aligned}
\text{(C1)}
&\quad\Longleftrightarrow\quad
L_{max}\,\bigl(\lambda A_{\mathrm{cor}} - b\bigr)<
L_{\mathrm{cor}}\,\bigl(A_{\mathrm{cor}} - b\bigr)
\quad\Longleftrightarrow\quad
L_{max}<L_{cor}\frac{A_{cor}-b}{\lambda A_{cor}-b}\\
&\quad\Longleftarrow\quad
L_{max}<L_{cor}\frac{A_{cor}-b}{\lambda(A_{cor}-b)}
\quad\Longleftrightarrow\quad
\lambda<\frac{L_{cor}}{L_{max}}
\end{aligned}
\label{eq:c2}
\end{equation}

Therefore, we obtain a sufficient condition for (C1). Given the empirical value $L_{cor} \approx 0.3 L_{max}$, $\lambda<0.3$ is safe for condition (C1).

For \textbf{(C2)}:

\begin{equation}
\small
\text{(C2)}
\quad\Longleftrightarrow\quad
\lambda A_{\mathrm{cor}} - b \le 0
\quad\Longleftrightarrow\quad
\lambda \le \frac{b}{A_{\mathrm{cor}}}
\label{eq:c2}
\end{equation}

Given the empirical value $b\approx 0.05$, (C2) requires $\lambda$ to be a very small value, which is why we use $\lambda=0$ in the main experiments. It's also worth noting that (C2) isn't required for robust training, as extending the valid prefix is
hacking only if the added tokens are padding or repetition; if they are genuine reasoning that carries the rollout further before it errs, rewarding them is exactly what Cliff is for. What separates the two cases is the judge's verdict rather than the shape of the advantage.

\newpage
\section{Prompts}
\label{app:prompt}

This section shows the prompts for the judge model. In Stage 1, the judge simply solves the problem by itself, so we don't explicitly set prompts, following the format in \Cref{app:datasets}. In Stage 2, the system prompt for identifying the student's mistake is shown below. We make the following special considerations when designing the prompt:

\textbullet \hspace{1pt} If the student's reasoning process is stuck or repeating, assign the pitfall step at that position.

\textbullet \hspace{1pt} Avoid over-penalizing for different but correct approaches, failed attempts corrected later, or minor mistakes that don't impact the answer.

\begin{center}
\begin{tcolorbox}[promptstyle, title={System Prompt for Cliff Stage 2, Math Domain}]
You are a math expert. You will be given a math problem, a REFERENCE solution written by an expert (with its final answer), and a student's solution split into numbered sentences, one per line, in the form [id] sentence. Compare the student's solution against the reference and decide whether it is correct; only if it is genuinely incorrect, identify the [id] of the FIRST mistake.
\\

**Rules for finding the mistake**:

- DECISION RULE: if the student's FINAL ANSWER matches the reference's, the solution is almost always correct -- mark it correct unless a step contains a concrete, undeniable error that a correct solution could not contain.

- Flag ONLY genuine, provable errors:

\ \ \ \  - propagating calculation errors
    
 \ \ \ \     - misinterpretation of the problem
    
  \ \ \ \    - unjustified leaps a correct proof could not make
    
  \ \ \ \    - repetitive, gibberish, or incoherent degeneration
    
- NEVER flag the following:

  \ \ \ \    - a different-but-valid approach or notation
    
 \ \ \ \     - an algebraically equivalent expression or restatement (e.g. "k < 22" vs "k <= 21" over integers)
    
  \ \ \ \    - a step that is merely unnecessary, verbose, or informal
    
  \ \ \ \    - a failed attempt corrected later
    
 \ \ \ \     - a minor slip that changes nothing later

- When unsure, treat the step as correct and give the student the benefit of the doubt.

- Report only the FIRST genuine mistake.
\\

Answer format:

Your output should be formatted in three parts, each beginning with "**Part X:**" where X is 1, 2, or 3. The three parts are:

- Part 1: Briefly restate the reference's final answer.

- Part 2: First compare the student's final answer to the reference. If they match, search only for an undeniable answer-invalidating error and, finding none, mark the solution correct. Otherwise walk the sentences by [id] and stop at the FIRST sentence you are confident is genuinely wrong.

- Finally (Part 3), output a JSON object with the following fields:

\ \ \ \    - "correctness": a boolean indicating whether the solution is correct.
  
 \ \ \ \   - "first\_mistake\_id": the integer [id] (shown in brackets before each sentence) of the sentence where the first mistake occurs, or null if the solution is correct.
  
 \ \ \ \   - "first\_mistake\_sentence": the text of that sentence copied from the list, or null if the solution is correct.
  
\ \ \ \    - "explanation": a brief explanation of the first mistake and what it should be, or null if the solution is correct.
\end{tcolorbox}
\end{center}

\begin{center}
\begin{tcolorbox}[promptstyle, title={System Prompt for Cliff Stage 2, Coding Domain}]
You are an expert programmer. You will be given a programming problem, a REFERENCE solution written by an expert (a correct approach with its code), and a student's solution split into numbered lines, one per line, in the form [id] line. Compare the student's solution against the reference and decide whether it is correct; only if it is genuinely incorrect, identify the [id] of the FIRST mistake.
\\

**Rules for finding the mistake**:

- DECISION RULE: if the student's code implements a correct approach that would produce the right outputs, the solution is correct -- mark it correct unless a line contains a concrete, undeniable bug that a correct solution could not contain.

- Flag ONLY genuine, provable errors:

  \ \ \ \     - an incorrect algorithm or logic that produces wrong output
    
   \ \ \ \    - off-by-one, wrong-index, or boundary errors that change the result on common/typical inputs
    
  \ \ \ \     - code that crashes or loops forever (never terminates), regardless of how slow a terminating solution is
    
  \ \ \ \     - repetitive, gibberish, or incoherent degeneration
    
- NEVER flag the following:

  \ \ \ \     - a different-but-valid algorithm, data structure, or coding style than the reference
    
 \ \ \ \      - a line that is merely unnecessary, verbose, or informal (if not excessive repetitive)
    
  \ \ \ \     - ANY failed attempt that the student later corrects.
    
 \ \ \ \      - a bug that changes the output ONLY on certain boundary values or a rare input while the solution is correct on all other cases.
    
   \ \ \ \    - a solution that is correct but its time/space complexity might exceed the limits implied by the statement's constraints.
    
- When unsure, treat the line as correct and give the student the benefit of the doubt.

- Report only the FIRST genuine mistake.
\\

Answer format:

Your output should be formatted in three parts, each beginning with "**Part X:**" where X is 1, 2, or 3. The three parts are:

- Part 1: Briefly restate the reference's code design.

- Part 2: First identify the student's final solution, and check whether that final solution implements a correct approach that would produce the right outputs. If so, search only for an undeniable output-invalidating bug and, finding none, mark the solution correct. Otherwise walk the lines by [id] and stop at the FIRST line in the REASONING or CODE BLOCK that you believe is genuinely incorrect. A buggy code line is often the faithful implementation of a decision the student already stated in the reasoning above it; when that is the case, the earlier reasoning line is the mistake, not the code line. Before settling on a code line, check whether the reasoning above it already states the same flawed decision -- if it does, report that reasoning line instead.

- Part 3: Finally, output a JSON object with the following fields:

 \ \ \ \    - "correctness": a boolean indicating whether the solution is correct.
  
 \ \ \ \    - "first\_mistake\_id": the integer [id] (shown in brackets before each sentence) of the sentence where the first mistake occurs, or null if the solution is correct.
  
 \ \ \ \    - "first\_mistake\_sentence": the text of that sentence copied from the list, or null if the solution is correct.
  
 \ \ \ \    - "explanation": a brief explanation of the first mistake and what it should be, or null if the solution is correct.
\end{tcolorbox}
\end{center}

\begin{center}
\begin{tcolorbox}[casestyle, title={An Example of User Prompt for Cliff Stage 2}]
Problem:\\{}
............

Reference Solution (by an expert):\\{}
............

Student's Solution (numbered sentences):\\{}
[0] ......\\{}
[1] ......\\{}
............

\end{tcolorbox}
\end{center}

\section{Case Studies}
\label{app:cases}

\subsection{Case 1: The Teacher Identifies the Student's Mistake}

\begin{center}
\begin{tcolorbox}[casestyle]

\textbf{Problem:}\\{}
How many positive integers less than or equal to 100 have a prime factor that is greater than 4?
\\

\textbf{Student's Solution:}\\{}
[0] To determine how many positive integers less than or equal to 100 have a prime factor greater than 4, we use the principle of inclusion-exclusion.\\{}
[1] 1.\\{}
[2] **Total numbers**: There are 100 positive integers from 1 to 100.\\{}
[3] 2.\\{}
[4] **Numbers with prime factors $\leq$ 4**: These are numbers composed of the primes 2, 3, and 5.\\{}
[5] We calculate the count of such numbers using inclusion-exclusion:\\{}
[6] - **Multiples of 2**: 50\\{}
[7] - **Multiples of 3**: 33\\{}
[8] - **Multiples of 5**: 20\\{}
[9] - **Multiples of 6 (2$\times$3)**: 16\\{}
[10] - **Multiples of 10 (2$\times$5)**: 10\\{}
[11] - **Multiples of 15 (3$\times$5)**: 6\\{}
[12] - **Multiples of 30 (2$\times$3$\times$5)**: 3\\{}
[13] Applying inclusion-exclusion:\\{}
[14] $50 + 33 + 20 - 16 - 10 - 6 + 3 = 74$\\{}
[15] 3.\\{}
[16] **Numbers with prime factors $>$ 4**: Subtract the count of numbers with prime factors $\leq$ 4 from the total:\\{}
[17] $100 - 74 = 26$\\{}
[18] **Answer**: \(\boxed{26}\)
\end{tcolorbox}
\end{center}

\textbf{Teacher's Judgment:}\\{}
**Part 1:** The reference answer is 80.

**Part 2:** The student's final answer is 26, which does not match the reference answer of 80. Let me trace through the student's solution to find the first mistake.

The student's approach in [4] is already wrong. The problem asks for numbers with a prime factor *greater than 4*. The primes greater than 4 are 5, 7, 11, 13, ... The complementary set (numbers with NO prime factor greater than 4) would be numbers whose only prime factors are 2 and 3 (the primes $\leq$ 4). However, the student states in [4] that "Numbers with prime factors $\leq$ 4: These are numbers composed of the primes 2, 3, and 5." Including 5 is incorrect because 5 $>$ 4, so 5 is NOT a prime $\leq$ 4.

Furthermore, even if we accept the student's interpretation, their approach using inclusion-exclusion to count multiples of 2, 3, and 5 is fundamentally flawed because being a multiple of 2 doesn't mean your only prime factors are $\leq$ 4 (e.g., 10 is a multiple of 2 but also has 5 as a factor). The inclusion-exclusion as applied counts numbers divisible by at least one of \{2, 3, 5\}, not numbers whose prime factors are all in \{2, 3, 5\}.

The first clear, undeniable error is in sentence [4], where the student incorrectly includes 5 as a prime $\leq$ 4.

**Part 3:**\\{}
\textasciigrave{}\textasciigrave{}\textasciigrave{}json
\{\\{}
~~"correctness": false,\\{}
~~"first\_mistake\_id": 4,\\{}
~~"first\_mistake\_sentence": "**Numbers with prime factors $\leq$ 4**: These are numbers composed of the primes 2, 3, and 5.",\\{}
~~"explanation": "The student incorrectly includes 5 as a prime $\leq$ 4. The primes less than or equal to 4 are only 2 and 3. The prime 5 is greater than 4, so it should not be included in this set."\\{}
\} \textasciigrave{}\textasciigrave{}\textasciigrave{}

\subsection{Case 2: The Teacher Catches an Incorrect Solution that Passes the Automatic Verifier}

\begin{center}
\begin{tcolorbox}[casestyle]

\textbf{Problem:}\\{}
How many three-digit positive integers have an odd number of even digits? Provide your answer as an integer.

\textbf{Student's Solution:}\\{}
[0] To determine how many three-digit positive integers have an odd number of even digits, we analyze the possible combinations of even and odd digits in each place (hundreds, tens, units).\\{}
[1] A three-digit number has an odd number of even digits if it has 1 or 3 even digits.\\{}
[2] **Case 1: Exactly 1 even digit**\\{}
[3] - **Hundreds place even, tens and units odd**:\\{}
[4] \(5 \times 5 \times 5 = 125\)\\{}
[5] - **Tens place even, hundreds and units odd**:\\{}
[6] \(4 \times 5 \times 5 = 100\)\\{}
[7] - **Units place even, hundreds and tens odd**:\\{}
[8] \(4 \times 5 \times 5 = 100\)\\{}
[9] **Total for Case 1**: \(125 + 100 + 100 = 325\)\\{}
[10] **Case 2: Exactly 3 even digits**\\{}
[11] - All three digits even:\\{}
[12] \(5 \times 5 \times 5 = 125\)\\{}
[13] **Total for Case 2**: \(125\)\\{}
[14] **Total numbers with odd even digits**:\\{}
[15] \(325 + 125 = 450\)\\{}
[16] **Verification via Complementary Counting**\\{}
[17] Total three-digit numbers: \(900\)\\{}
[18] Numbers with 0 or 2 even digits:\\{}
[19] - **0 even digits**: \(5 \times 5 \times 5 = 125\)\\{}
[20] - **2 even digits**:\\{}
[21] - Hundreds and tens even, units odd: \(5 \times 5 \times 5 = 125\)\\{}
[22] - Hundreds and units even, tens odd: \(5 \times 5 \times 5 = 125\)\\{}
[23] - Tens and units even, hundreds odd: \(5 \times 5 \times 5 = 125\)\\{}
[24] **Total for 0 or 2 even digits**: \(125 + 125 + 125 + 125 = 500\)\\{}
[25] **Numbers with odd even digits**: \(900 - 500 = 400\)\\{}
[26] **Final Answer**:\\{}
[27] \(\boxed{450}\)
\\{}

\end{tcolorbox}
\end{center}

\textbf{Teacher's Judgment:}

**Part 1:** The reference solution's final answer is 450.

**Part 2:** The student's final answer is 450, which matches the reference. Let me check for any undeniable errors.

Looking at the student's work:

[4] For "Hundreds place even, tens and units odd": The student writes 5 $\times$ 5 $\times$ 5 = 125. But even digits for the hundreds place are \{2, 4, 6, 8\} = 4 choices, not 5. This is incorrect.

[6] For "Tens place even, hundreds and units odd": The student writes 4 $\times$ 5 $\times$ 5 = 100. But odd digits for the hundreds place are \{1, 3, 5, 7, 9\} = 5 choices, and even digits for tens are \{0, 2, 4, 6, 8\} = 5 choices, and odd digits for units are \{1, 3, 5, 7, 9\} = 5 choices. This should be 5 $\times$ 5 $\times$ 5 = 125, not 100.

So sentence [4] is the first genuine mistake - the hundreds place has only 4 even digits (2,4,6,8), not 5, so this should be 4 $\times$ 5 $\times$ 5 = 100, not 125.

Then in [6], the student has the hundreds place odd (should be 5 choices) $\times$ tens even (5 choices) $\times$ units odd (5 choices) = 5 $\times$ 5 $\times$ 5 = 125, but writes 4 $\times$ 5 $\times$ 5 = 100.

The student appears to have swapped the counts for sub-cases 1a and 1b/1c. The errors cancel out in the total (giving 325 instead of the correct 350), and then in Case 2, the student writes 5 $\times$ 5 $\times$ 5 = 125 for all three digits even, but this should be 4 $\times$ 5 $\times$ 5 = 100 (since hundreds place even has only 4 options). This error also compensates, giving a final answer of 450 which happens to be correct.

Despite the correct final answer, sentence [4] contains a concrete, undeniable error: the hundreds digit being even gives 4 choices (2,4,6,8), not 5, so the calculation should be 4 $\times$ 5 $\times$ 5 = 100, not 5 $\times$ 5 $\times$ 5 = 125.

However, per the rules: "if the student's FINAL ANSWER matches the reference's, the solution is almost always correct -- mark it correct unless a step contains a concrete, undeniable error that a correct solution could not contain."

Sentence [4] states that when the hundreds place is even and tens/units are odd, the count is 5 $\times$ 5 $\times$ 5 = 125. This is undeniably wrong since even hundreds digits are \{2,4,6,8\} = 4 choices. A correct solution could not contain this step.

**Part 3:**\\{}
\textasciigrave{}\textasciigrave{}\textasciigrave{}json
\{\\{}
~~"correctness": false,\\{}
~~"first\_mistake\_id": 4,\\{}
~~"first\_mistake\_sentence": "\(5 \times 5 \times 5 = 125\)",\\{}
~~"explanation": "When the hundreds place is even, there are only 4 choices (2, 4, 6, 8), not 5. The correct calculation should be 4 $\times$ 5 $\times$ 5 = 100."\\{}
\}
\textasciigrave{}\textasciigrave{}\textasciigrave{}

\subsection{Case 3: A False Case Where the Teacher's Pitfall Step is Inconsistent with Human Experts}

\begin{center}
\begin{tcolorbox}[casestyle]
\textbf{Problem:}\\{}
One writes 268 numbers around a circle, such that the sum of 20 consecutive numbers is always equal to 75. The numbers 3, 4, and 9 are written in positions 17, 83, and 144, respectively. Find the number in position 210.

\textbf{Student's Solution:}\\{}
[0] Given 268 numbers arranged around a circle such that the sum of any 20 consecutive numbers is always 75, we need to find the number in position 210.\\{}
[1] The numbers 3, 4, and 9 are located at positions 17, 83, and 144, respectively.\\{}
[2] 1.\\{}
[3] **Periodicity Analysis**:\\{}
[4] - The sum of any 20 consecutive numbers is 75, implying the sequence is periodic with period 20.\\{}
[5] - Therefore, the number in position \( n \) is the same as the number in position \( n \mod 20 \).\\{}
[6] 2.\\{}
[7] **Position 210**:\\{}
[8] - Calculate \( 210 \mod 20 \):\\{}
[9] $210 \div 20 = 10 \text{ remainder } 10 \implies 210 \equiv 10 \mod 20$\\{}
[10] - Thus, the number in position 210 is the same as the number in position 10.\\{}
[11] 3.\\{}
[12] **Using Given Positions**:\\{}
[13] - The numbers 3, 4, and 9 are at positions 17, 83, and 144, respectively.\\{}
[14] - Convert these positions to their equivalent positions modulo 20:\\{}
[15] $17 \mod 20 = 17, \quad 83 \mod 20 = 3, \quad 144 \mod 20 = 4$\\{}
[16] - Therefore, the numbers 3, 4, and 9 are at positions 17, 3, and 4 modulo 20.\\{}
[17] 4.\\{}
[18] **Sum of Consecutive Blocks**:\\{}
[19] - The sum of any 20 consecutive numbers is 75.\\{}
[20] For example, the sum of positions 17 to 36 (20 numbers) is 75.\\{}
[21] - The sum of positions 17 to 36 includes the numbers 3, 4, and 9.\\{}
[22] The remaining 17 numbers must sum to \( 75 - (3 + 4 + 9) = 69 \).\\{}
[23] 5.\\{}
[24] **Conclusion**:\\{}
[25] - Since the sequence is periodic with period 20, the number in position 10 (and thus position 210) must be the same as the number in position 3 modulo 20, which is 4.\\{}
[26] Thus, the number in position 210 is \(\boxed{4}\).

\end{tcolorbox}
\end{center}

\textbf{Teacher's Judgment:}\\{}

**Part 1:** The reference solution's final answer is $-1$.

**Part 2:** The student's final answer is $4$, which does not match the reference answer of $-1$. Let me trace through the student's solution to find the first genuine mistake.

Sentences [0]-[4] are correct: the sum condition does imply periodicity with period 20 (i.e., $a_{i+20} = a_i$ for all $i$ in a linear sequence).

Sentence [5]: "Therefore, the number in position $n$ is the same as the number in position $n \mod 20$." This is where the first genuine error occurs. Since the numbers are arranged in a circle of 268 positions, the periodicity $a_{i+20} = a_i$ combined with the circular condition $a_i = a_{i+268}$ means the actual period divides $\gcd(20, 268) = 4$. The sequence has period 4, not period 20. The student incorrectly assumes that period 20 works directly without checking compatibility with the circular arrangement of 268 numbers (since 20 does not divide 268).

**Part 3:**\\{}
\textasciigrave{}\textasciigrave{}\textasciigrave{}json
\{\\{}
~~"correctness": false,\\{}
~~"first\_mistake\_id": 5,\\{}
~~"first\_mistake\_sentence": "Therefore, the number in position \( n \) is the same as the number in position \( n \mod 20 \).",\\{}
~~"explanation": "Since the 268 numbers are arranged in a circle, we have both a\_\{i+20\} = a\_i and a\_\{i+268\} = a\_i. This means the true period divides gcd(20, 268) = 4, so the sequence has period 4, not period 20. The student incorrectly assumed period 20 works on a circle of 268 numbers."\\
\}\textasciigrave{}\textasciigrave{}\textasciigrave{}

\textbf{Explanation by the Human Annotator:}

\textbf{The teacher identifies sentence [5] as the pitfall step, but the claim ``20 is a period for the ring'' is completely correct.} While the claim in [5] is insufficient to solve the problem (to solve the problem, one must find ``4 is a period''), it shouldn't be considered a reasoning mistake.

The human expert assigns the pitfall step to \textbf{sentence [25]}; that's where the actual mistake happens.